\documentclass[10pt,twocolumn,letterpaper]{article}

\long\def\comment#1{}

\usepackage{cvpr}
\usepackage{times}
\usepackage{epsfig}
\usepackage{graphicx}
\usepackage{amsmath}
\usepackage{amssymb}
\usepackage[listofformat=parens, justification=centering, subrefformat=subparens]{subfig}

\DeclareMathOperator*{\argmin}{arg\,min}
\renewcommand\cap[3]{\caption[#2]{\label{#1}\textsc{#2}. \small\textit{#3}}}
\usepackage[pagebackref=true,breaklinks=true,letterpaper=true,colorlinks,bookmarks=false]{hyperref}

\renewcommand\cap[3]{\caption[#2]{\label{#1}\textsc{#2}. \small\textit{#3}}}
\graphicspath{{./images/}}

\makeatletter
\renewcommand\section{\@startsection {section}{1}{\z@}
                                   {-1.9ex \@plus -2ex \@minus -.2ex}
                                   {.6ex \@plus.2ex}
                                   {\normalfont\Large\bfseries}}
\renewcommand\subsection{\@startsection{subsection}{2}{\z@}
                                   {-1.3ex \@plus -.8ex \@minus -.2ex}
                                   {.4ex \@plus.2ex}
	                           {\normalfont\large\bfseries}}
\renewcommand\subsubsection{\@startsection{subsubsection}{3}{\z@}
                                     {-.5ex\@plus -.2ex \@minus -.2ex}
                                     {.2ex \@plus .2ex}
                                     {\normalfont\large\bfseries}}

\def\tightmath{
\abovedisplayskip=4pt plus 2pt minus 1pt 
\abovedisplayshortskip=2pt plus 1pt minus 1pt 
\belowdisplayskip=4pt plus 2pt minus 1pt 
\belowdisplayshortskip=2pt plus 1pt minus 1pt }
\def\crushmath{
\abovedisplayskip=1pt plus 1pt minus 2pt 
\abovedisplayshortskip=1pt plus 1pt minus 2pt 
\belowdisplayskip=1pt plus 1pt minus 2pt 
\belowdisplayshortskip=1pt plus 1pt minus 2pt }
\tightmath
\crushmath

\makeatother

\cvprfinalcopy % *** Uncomment this line for the final submission

\def\cvprPaperID{18} % *** Enter the CVPR Paper ID here

\ifcvprfinal\fi
\begin{document}
\crushmath
%%%%%%%%% TITLE
\title{Adversarial Diversity and Hard Positive Generation}

\author{Andras Rozsa, Ethan M. Rudd, and Terrance E. Boult\thanks{This work supported in part by NSF\#1320956 RI: Small: Open Vision}\\
University of Colorado at Colorado Springs\\
Vision and Security Technology (VAST) Lab\\
\{arozsa,erudd,tboult\}@vast.uccs.edu}

%\author{Andras Rozsa\\
%\and

\maketitle
\thispagestyle{empty}
\vspace*{-2ex}
%%%%%%%%% ABSTRACT
\begin{abstract}
\vspace*{-2ex}
State-of-the-art deep neural networks suffer from a fundamental problem -- they misclassify adversarial examples formed by applying small perturbations to inputs. In this paper, we present a new psychometric perceptual adversarial similarity score (PASS) measure for quantifying adversarial images, introduce the notion of \textit{hard positive generation}, and use a diverse set of adversarial perturbations -- not just the closest ones -- for data augmentation. We introduce a novel hot/cold approach for adversarial example generation, which provides multiple possible adversarial perturbations for every single image. The perturbations generated by our novel approach often correspond to semantically meaningful image structures, and allow greater flexibility to scale perturbation-amplitudes, which yields an increased diversity of adversarial images. We present adversarial images on several network topologies and datasets, including LeNet on the MNIST dataset, and GoogLeNet and ResidualNet on the ImageNet dataset. Finally, we demonstrate on LeNet and GoogLeNet that fine-tuning with a diverse set of hard positives improves the robustness of these networks compared to training with prior methods of generating adversarial images.
\end{abstract}

% Previous..
%State-of-the-art deep neural networks suffer from a fundamental problem -- they misclassify adversarial examples formed by applying small perturbations to inputs. In this paper, we present a new psychometric \textit{perceptual adversarial similarity score} (PASS) measure for quantifying adversarial images, and introduce the notion of \textit{hard positive generation}, using a diverse set of adversarial perturbations -- not just the closest one -- to augment the data. We introduce a novel \textit{hot/cold} approach to hard positive generation, which unlike most previous approaches, yields \textit{many} directions of adversarial perturbations. Our \textit{hot/cold} approach can be used to create images with better PASS scores than current approaches, with many perturbations corresponding to semantically meaningful image structures, allowing greater flexibility to scale perturbation amplitudes to yield a great diversity of adversarial images.   We demonstrate examples of adversarial and hard positive images on several network topologies and datasets, including LeNet on the MNIST dataset and GoogLeNets and ResidualNets on the ImageNet dataset, and show that fine-tuning with a diverse set of hard positives improves the robustness of these networks compared to training with prior methods of generating adversarial images.

%\vspace{-2em}
\section{Introduction}
\label{Intro}

\begin{figure}[!t]
\begin{center}
  \centerline{\includegraphics[width=.95\columnwidth]{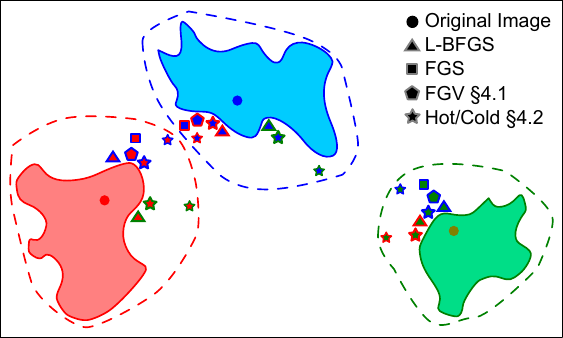}}
%\vspace*{-0.15in}
\cap{fig:general}{Adversarials and Hard Positives}{This paper demonstrates how to generate a much more diverse set of adversarial examples than the existing L-BFGS~\cite{szegedy2013intriguing} or fast gradient sign (FGS)~\cite{goodfellow2014explaining} methods. Via a range of perturbation amplitudes along the learnt adversarial directions -- not just the closest adversarial sample -- we can generate hard positives to fine-tune the class definitions, thereby extending previously overfit decision boundaries to improve both accuracy and robustness. The extended decision boundaries are represented by dashed lines. This simplified schematic uses shapes to depict different types of hard positive examples. Inner colors depict the original class, while outer colors depict the classification by the base network. For better visualization we show only one input image with corresponding adversarial/hard positive examples for each class.}
\end{center}
\vspace*{-0.25in}
\end{figure}

Deep neural networks are powerful learning models which have been successfully applied to vision, speech and many other tasks~\cite{he2015deep,krizhevsky2012imagenet,Simonyan15,socher2011parsing,jia2014caffe,szegedy2015going,clark2015training,chen2015learning}. Szegedy et al. \cite{szegedy2013intriguing} showed that several machine learning models, including state-of-the-art deep neural networks, misclassify small non-random perturbations of correctly classified images. In many cases, these misclassifications are made with high confidence. Szegedy et al. \cite{szegedy2013intriguing} dubbed these perturbed misclassified samples \emph{adversarial examples}. In order to  generalize well, deep neural networks are expected to be robust to moderate perturbations to their inputs. Thus adversarial examples, with only small perturbations, are problematic.

A deeper problem is that the generated adversarial images are relatively portable across different neural networks, which  means that they are consistently misclassified by models of similar network architectures trained on varying training data, with different hyperparameters, or even different numbers of layers or types of activations~\cite{szegedy2013intriguing}. While one classification error is a simple practical problem, these highly unexpected recognition errors suggest a more fundamental problem. Namely, the combinations of training samples and algorithms that we use to train our networks are not sufficient. 
 
At their core, adversarial examples are nothing more than perturbed versions of ordinary examples that cause unexpected recognition mistakes in networks. This suggests that the adversarial problem can be addressed by finding ways to efficiently augment the training set with representative adversarial examples which increase the diversity and thus generalization of the training set. Enhancing the training set is a common technique in machine learning and has been applied by others to deep neural networks. For example, Bengio et al. \cite{bengio2011deep} demonstrated that deep networks benefit even more from out-of-distribution examples including perturbed versions of the original training examples.

In this paper, we make the following contributions:
\begin{enumerate}
  \vspace*{-0.08in}
  \itemsep -3pt
  \parsep -3pt
\item We introduce a new measure to quantify \textit{adversarial images} using a novel \textit{Perceptual Adversarial Similarity Score} (PASS), which is more consistent with human perception than prior $L_p$ norm measurements.

\item We introduce a new approach which is capable of generating large numbers of diverse adversarial images.

\item While researchers have focused on generating the closest adversarial images, we argue that to augment training sets and thereby improve accuracy and robustness of learning models, it is better to use \textit{hard positives} formed by non-minimal perturbations. We demonstrate that these non-minimal hard positives are more effective for training than existing adversarial models.
\end{enumerate}

%We hypothesize that augmenting the training set with adversarial images improves robustness and classification accuracy, not by choosing the closest adversarial images (high PASS score), but by training with adversarial images having lower PASS values. 
%Finally, we perform experiments which validate our hypotheses and suggest that training on adversarial samples can improve both robustness and overall classification performance of deep neural networks.
%First, we formalize the term \textit{adversarial image} and introduce the \textit{Perceptual Adversarial Similarity Score} (PASS), a new metric which measures ``adversarialness'' by calculating similarities of original and perturbed images. Second, we demonstrate several new methods which are capable of generating a vast amount of diverse adversarial images. We hypothesize that augmenting the training set with adversarial images improves robustness and classification accuracy, not by choosing the closest adversarial images (high PASS score), but by training with adversarial images having lower PASS values. %but by training with adversarial images with intermediate PASS values and a greater diversity of adversarial data. 
%Finally, we perform experiments which validate our hypotheses and suggest that training on adversarial samples can improve both robustness and overall classification performance of deep neural networks.

\section{Related Work}

While not called \textit{adversarial examples}, the oldest work related to our research is the use of perturbations and/or hard negative mining for training. Introducing perturbations to inputs to improve learning machines is a long standing method in machine learning, e.g., it was used when the MNIST dataset \cite{lecun1998mnist} was introduced and put forward as a training methodology in \cite{simard2003best}. Research by Loosli et al. \cite{loosli2007training} took perturbation-enhanced training to a new level and developed an open source tool called InfiMNIST \cite{loosli2007training} that produces MNIST examples by applying small affine transformations and noise to the original images.

In detection problems, where the number of negatives can be enormous, a related technique is \textit{hard negative mining} in which naturally occurring but ``hard'' and hence informative examples are used to improve training, e.g.,~\cite{felzenszwalb2008discriminatively}.

Baluja et al.~\cite{baluja2015virtues} proposed a related approach which generates perturbations and applies them to the inputs, observing how multiple trained networks respond to each input. Their method applies small affine image transformations without having any knowledge about the inner states of the networks. They used peer networks as a control-mechanism to filter out radical perturbations and to identify which perturbations are useful for retraining. However, random perturbations can be an inefficient technique to generate good training data, since a good system will correctly classify the vast majority of such inputs. We generated one million new examples with InfiMNIST and tested them on three differently trained LeNet \cite{lecun1995learning} networks to identify adversarial examples among these images. The results show that the proportion of adversarial examples in the InfiMNIST dataset is very small -- 2.199$\pm$0.132\% -- which means that finding the small affine transformations that form adversarial examples via this method is computationally non-trivial. Although capable of finding high-value training examples, these types of \textit{guess and check} approaches, which perform random perturbations and determine if the results are misclassified, can be prohibitively expensive.

The idea of using optimization and internal network state to find adversarial examples in machine learning models was introduced in \cite{szegedy2013intriguing}. The authors also demonstrated the first method to reliably find those perturbations via small adjustments to pixel values. The approach relies on a box-constrained optimization technique (L-BFGS) which, starting from a randomly chosen direction, aims to find the smallest perturbation in the input space that causes the perturbed image to be classified as a predefined target label. Szegedy et al.~\cite{szegedy2013intriguing} performed several experiments on a few varying networks and datasets, including MNIST \cite{lecun1998mnist}, and demonstrated that the same adversarial example is often misclassified by different networks.

In \cite{goodfellow2014explaining}, Goodfellow et al. presented a simpler and computationally cheaper algorithm to produce small perturbations causing unexpected recognition errors. Their approach -- the fast gradient sign (FGS) method -- creates perturbations by using the sign of the gradient of loss with respect to the input, and the required gradient can be effectively calculated using backpropagation. Experiments in the paper demonstrated that FGS reliably causes a wide variety of learning models to misclassify their perturbed inputs. It is important to note that while both L-BFGS and FGS use gradient information, FGS is much faster because the gradient is used only once as an explicit direction for a line-search, as opposed to L-BFGS, which performs many gradient computations. Stepping in the direction of the sign of the gradient of loss with respect to the input image continuously reduces the classification score of the original class until another class obtains a higher score. Assuming that the original classification was correct, this causes a classification error. The resulting perturbation images look like dense random noise. Rather than producing images for training, the paper suggests using an improved objective function which directly incorporates the sign of the gradient of loss. That model reduces the average classification error rate from 0.99\% to 0.782\% on MNIST.

A fundamentally different approach from~\cite{szegedy2013intriguing,goodfellow2014explaining} was proposed by Sabour et al.\cite{sabour2015adversarial}. The approach seeks to find not only adversarial images that are misclassified; it also seeks bounded error inputs that have internal representations which are closest to \textit{guide images}. In their work, the authors demonstrate that adversarial examples can be produced which are not only incorrectly classified at the output layer, but are also close to any specified internal representations of the network. They use L-BFGS to find the adversarial images which mimic the internal representations of targeted images, also demonstrating that the adversarial problem is more complex than just mapping output errors.  

In all research that uses explicitly optimized adversarial images to improve networks, the authors aim to synthesize the perturbations of smallest difference (by some measure) that cause misclassification. By contrast, we further amplify adversarial perturbations over a diverse range to obtain additional \textit{hard positives} to retrain our networks.

\section{PASS}

\begin{figure*}[t!]
\begin{center}
\centering\subfloat[][\label{sfig:a}\centering Fast Gradient Sign \par \textit{PASS}=$0.727$, $L_2$=$0.990$, $L_\infty$=$13$]{\includegraphics[]{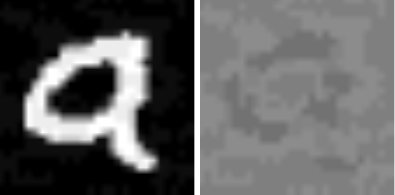}}
\hfill
\centering\subfloat[][\label{sfig:b}\centering Fast Gradient Value \par \textit{PASS}=$0.979$, $L_2$=$0.196$, $L_\infty$=$22$]{\includegraphics[]{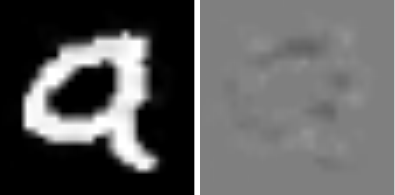}}
\hfill
\centering\subfloat[][\label{sfig:c}\centering Hot/Cold - 1 \par \textit{PASS}=$0.977$, $L_2$=$0.198$, $L_\infty$=$22$]{\includegraphics[]{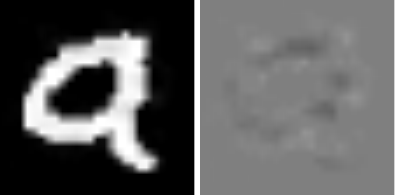}}
\hfill
\centering\subfloat[][\label{sfig:d}\centering Hot/Cold - 2 \par \textit{PASS}=$0.965$, $L_2$=$0.239$, $L_\infty$=$24$]{\includegraphics[]{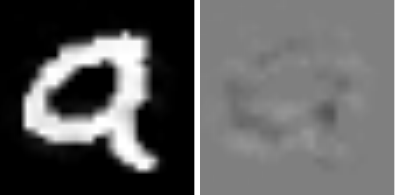}}

\cap{fig:first_gen}{Adversarials on LeNet/MNIST}{Adversarial examples and perturbations for an input image classified as digit 9 with metrics shown below images in form (\textit{PASS}, L$_2$ norm per pixel, L$_\infty$). (a) Fast Gradient Sign method: adversarial classified as 4 (b) Fast Gradient Value approach: adversarial classified as 4 (c) Hot/Cold approach with the most similar class: adversarial classified as 4 (d) Hot/Cold approach with the second most similar class: adversarial classified as 2. Perturbation-visualization: black=-1, white=+1, gray=no-change.}
\end{center}
\vspace*{-0.15in}
\end{figure*}

Different measures have been used in the literature to quantify adversarial images; commonly $L_p$ norms~\cite{szegedy2013intriguing,goodfellow2014explaining,baluja2015virtues}.
These works implicitly agree that \textit{adversarial images} are modified inputs which cause unexpected recognition errors in machine learning models, yet \textit{are correctly classified by humans}. Sabour et al. concluded that $L_p$ measures are not well matched to human perception \cite{sabour2015adversarial}.
Consistent with \cite{flynn2013image}, this suggests that adversarial images should be quantified  in terms of \textit{just noticeable difference}. However, a natural interpretation of \textit{imperceptible} should also allow many transformations, including small translations and rotations, which result in images that are perturbed to noticeable extents compared to their original counterparts, yet still appear to be \textit{plausible} samples of the same images -- samples that a network should not get wrong. Consider a biometric face verification system under attack: viewpoint variations result in noticeably different images that a human operator still perceives as a different view of the same input. If the distortions are too large, e.g., a face with reversed aspect ratio or a very noisy image, a human operator would likely notice a clear problem with the data even if he or she could still discern the identity of the person in the image.

To quantify the degree to which an image is adversarial, we therefore seek a psychometric measure which considers not only element-wise similarity but also \textit{plausibility} that the image in question is a different view of the same input. We perform this measurement as a two stage process: aligning the perturbed image with the original, then measuring similarity of the aligned images. Due to potential radiometric and noise differences between the images, simple correlation or feature-based alignments may not work very well. Instead, we maximize the \textit{enhanced correlation coefficient} (ECC)~\cite{evangelidis2008parametric} between the adversarial image $\tilde{x}$ and the original image $x$. Let $\psi(\tilde{x},x)$ be a homography transform from adversarial image $\tilde{x}$ to the original image $x$, with $3\times 3$ homography matrix ${\bf H}$. We optimize the objective
\begin{equation}
\label{eq:ecc_objective}
\argmin_{\bf H} \left|\left| \frac{\overline{x}}{||\overline{x}||} - \frac{\overline{\psi(\tilde{x},x)}}{||\overline{\psi(\tilde{x},x)}||} \right|\right|,
\end{equation}
where an overline ($\overline{\cdot}$) denotes the zero-mean version of an image. The $x$ minimizing Eq.~\ref{eq:ecc_objective} will maximize $ECC(\psi(\tilde{x},x),x)$.  

The second stage of our comparison measures similarity. Previous work~\cite{szegedy2013intriguing,sabour2015adversarial} has predominantly quantified the degree of \textit{adversarial} in terms of element-wise $L_2$ or $L_\infty$ distances. However, these norms are extremely sensitive to small geometric distortions and do not map well to psychophysical notions of similarity, even under non-geometric perturbations; even after alignment, a single pixel-wise difference along a strong edge is all it takes to maximize the $L_\infty$ distance between two very similar images.
%. As we see in Fig.~\subref{sfig:g}, even after alignment, a single pixel-wise difference along a strong edge is all it takes to maximize the $L_\infty$ distance between two very similar images.

\comment{
  Research  by~\cite{wang2004image} suggests  that the human visual system is most sensitive to changes in \textit{structural patterns} and developed the structural similarity (SSIM) index as an objective for lossy image compression. 
SSIM quantifies similarity of image pairs based on their structural and brightness differences. Given two images, $x$ and $y$, let $L(x,y)$, $C(x,y)$, and $S(x,y)$ be luminance, contrast, and structural measures, respectively defined as 
\begin{align*}
L(x,y) &= \left[\frac{2\mu_x\mu_y + C_1}{{\mu_x}^2+{\mu_y}^2 + C_1}\right]\\
C(x,y) &= \left[\frac{2\sigma_x\sigma_y + C_2}{{\sigma_x}^2+{\sigma_y}^2 + C_2} \right]\\
S(x,y) &= \left[\frac{\sigma_{xy} + C_3}{\sigma_x\sigma_y+C_3}\right].
\end{align*}
}
Research  by~\cite{wang2004image} suggests  that the human visual system is most sensitive to changes in \textit{structural patterns} and developed the structural similarity (SSIM) index as an objective for lossy image compression. 
SSIM quantifies similarity of image pairs based on structural and brightness differences. 

Given two images, $x$ and $y$, let $L(x,y)$, $C(x,y)$, and $S(x,y)$ be luminance, contrast, and structural measures, respectively defined as
\begin{align*}
L(x,y) &= \left[\frac{2\mu_x\mu_y + C_1}{{\mu_x}^2+{\mu_y}^2 + C_1}\right]\\
C(x,y) &= \left[\frac{2\sigma_x\sigma_y + C_2}{{\sigma_x}^2+{\sigma_y}^2 + C_2} \right]\\
S(x,y) &= \left[\frac{\sigma_{xy} + C_3}{\sigma_x\sigma_y+C_3}\right],
\end{align*}
where $\mu_x$, $\sigma_x$, and $\sigma_{xy}$
are weighted mean, variance, and covariance respectively and $C_i$'s are constants to prevent singularity. With these, the regional SSIM index (RSSIM) is defined as 
\begin{equation}
\textit{RSSIM}(x,y) = L(x,y)^\alpha C(x,y)^\beta S(x,y)^\gamma,
\end{equation}
where $\alpha$, $\beta$, and $\gamma$ are chosen to reflect relative importance of luminance, contrast, and structure respectively. 
For consistency with~\cite{wang2004image}, we set $\alpha = \beta = \gamma = 1$, and use an $11\times11$ kernel of $\sigma=1.5$ for weights.
SSIM is then obtained by taking the average of RSSIM over all pixels: for an $m$-pixel image, 
\begin{equation}
\textit{SSIM}(x,y) = \frac{1}{m} \sum_{n=1}^{m} \textit{RSSIM}(x_n,y_n).
\end{equation}
We combine the  photometric-invariant homography transform alignment with SSIM to define the \textit{perceptual adversarial similarity score} (PASS) between $\tilde{x}$ and $x$ as
\begin{equation}
\textit{PASS}(\tilde{x},x) = \textit{SSIM}(\psi(\tilde{x},x),x).
\end{equation}

PASS then serves as a similarity measure to quantify how \textit{adversarial} a misclassified image is. While previous works quantified adversarial in terms of some similarity or dissimilarity measure, both Szegedy et al.~\cite{szegedy2013intriguing} and Goodfellow et al.~\cite{goodfellow2014explaining} mention a constraint which they implicitly assume but do not explicitly define: namely, in order to be adversarial perturbations must be \textit{imperceptible}. Mathematical definitions purely in terms of $L_p$ norms do not operationally enforce such a constraint; the perturbation of minimum $L_p$ norm may be quite perceptible for certain images. Insofar as PASS serves as a psychometric, we can use it to make this constraint explicit. Let $y$ be the label of $x$, let $f$ be the classifier, and let $\tau \in [0,1]$ be a threshold on perceptible PASS values. Then an adversarial image is defined as
\begin{equation}
\label{eq:adv}
\argmin_{d(x,\tilde{x})} \tilde{x} :f (\tilde{x}) \neq y \text{ and } \textit{PASS}(x,\tilde{x}) \geq \tau,
\end{equation}
where $d(x,\tilde{x})$ is some dissimilarity measure, e.g., $1-\textit{PASS}(x,\tilde{x})$ or $||x-\tilde{x}||_p$ -- potentially constrained along the directions learnt by an adversarial generation algorithm. Note that the value of $\tau$ may vary depending on the network and the dataset, but for any fixed domain this gives a quantitative adversarial threshold. Hard positives are similarly constrained by a \textit{PASS} threshold, but need not be the samples of minimum dissimilarity.

%If one seeks a binary classifier for ``adversarial'', then
%\begin{equation}\label{eq:adv}
%adv(\tilde{x}{,x)} = \left\{\def\arraystretch{1.2}%                                                                                                                                 
%  \begin{array}{@{}c@{\:\:}l@{}}
%    {1} & \mathit{if} \textit{ PASS}{(\tilde{x},x) \geq \tau \: \mathrm{and} \: f(\tilde{x}) \neq y}\\
%    {0} & \mathrm{otherwise.}\\
%  \end{array}\right.
%\end{equation}
%As we shall see in our experiments, the ideal value of $\tau$ is application dependent, but for any fixed domain this gives a quantitative adversarial measure.

\section{Adversarial Example Generation}

In this section, we first provide an overview of the notations that we use. We then discuss an existing method for generating adversarial examples. Finally, we introduce a novel method for adversarial image and hard positive generation and discuss implementation details.

Let $\theta$ be the parameters of the model, $x \in [0,255]^m$ the $m$-pixel image of integer values applied as input to the network, $y$ the label of $x$, $J(\theta,x,y)$ the cost used to train the neural network, and $f$ be the learnt classifier that maps input images to a discrete set of $n$ labels. Let $B_l(\cdot)$ be the backpropagation operator defined in Sec.~\ref{sec:hack}.

For a given input image $x$ classified as $y$, our goal is to produce perturbation $\eta$ such that perturbed image $\tilde{x}=x+{\eta}$ is adversarial according to Eq.~\ref{eq:adv}. To generate hard positives, we simply scale $\eta$ by a constant $\geq 1$.

%We convert floating point images in $[0,1]$ to the integer domain $[0,255]$ and back as needed. 

\subsection{Fast Gradient Value}
\label{sec:fgv}

We commence our research using the fast gradient sign (FGS) method introduced in \cite{goodfellow2014explaining}, but seek greater adversarial diversity. An obvious extension of FGS is to consider a scaled version of the raw gradient of loss instead of using only the sign of the gradient.  We refer to this as fast gradient value (FGV), and we show that it produces notably different adversarial perturbations. Specifically, the direction determined by  
\begin{equation}\label{eq:gradient_of_loss}
\eta_{grad} = \nabla_xJ(\theta,x,y)
\end{equation}
does not ignore the differences in gradient magnitude between corresponding pixels as FGS does. The intuition here is that to effectively increase the loss with minimal distortion, pixels with larger gradients in terms of $L_1$ norm need to be changed more radically than others. Compared to FGS, this slight modification leads to different directions where significantly different adversarial examples exist. 

\begin{figure*}[t!]
\begin{center}
\centering\subfloat[][\label{face_gen:a}\centering BVLC-GoogLeNet: FGS \par \textit{PASS}=$0.341$, $L_2$=$0.162$, $L_\infty$=$21$]{\includegraphics[]{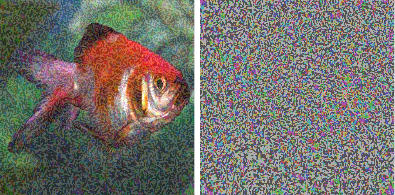}}
\hfill
\centering\subfloat[][\label{face_gen:b}\centering BVLC-GoogLeNet: FGV \par \textit{PASS}=$0.663$, $L_2$=$0.089$, $L_\infty$=$120$]{\includegraphics[]{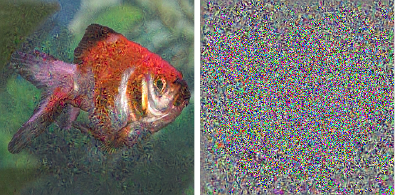}}
\hfill
\centering\subfloat[][\label{face_gen:c}\centering BVLC-GoogLeNet: HC-1 \par \textit{PASS}=$0.986$, $L_2$=$0.012$, $L_\infty$=$25$]{\includegraphics[]{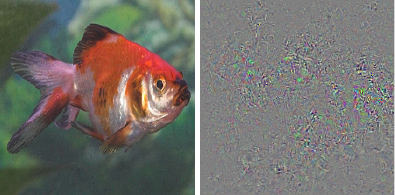}}
\hfill
\centering\subfloat[][\label{face_gen:d}\centering BVLC-GoogLeNet: HC-13 \par \textit{PASS}=$0.975$, $L_2$=$0.016$, $L_\infty$=$25$]{\includegraphics[]{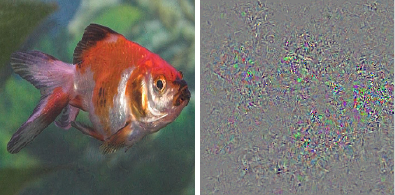}}
\\
\centering\subfloat[][\label{face_gen:e}\centering ResNet-152: FGS \par \textit{PASS}=$0.917$, $L_2$=$0.031$, $L_\infty$=$4$]{\includegraphics[]{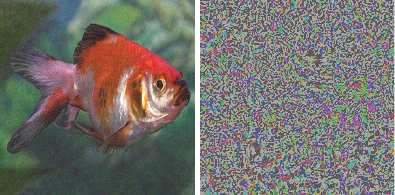}}
\hfill
\centering\subfloat[][\label{face_gen:f}\centering ResNet-152: FGV \par \textit{PASS}=$0.983$, $L_2$=$0.012$, $L_\infty$=$17$]{\includegraphics[]{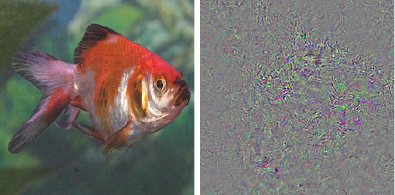}}
\hfill
\centering\subfloat[][\label{face_gen:g}\centering ResNet-152: HC-1 \par \textit{PASS}=$0.992$, $L_2$=$0.008$, $L_\infty$=$14$]{\includegraphics[]{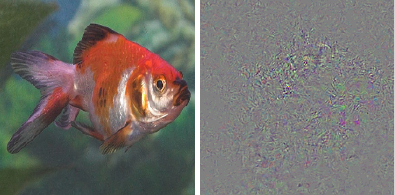}}
\hfill
\centering\subfloat[][\label{face_gen:h}\centering ResNet-152: HC-13 \par \textit{PASS}=$0.997$, $L_2$=$0.004$, $L_\infty$=$7$]{\includegraphics[]{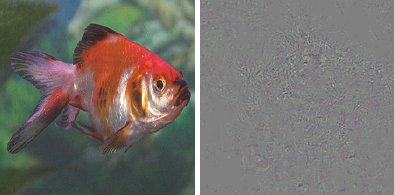}}

\cap{fig:face_gen}{Adversarial Images on ImageNet}{Adversarial examples and perturbations for a ``goldfish'' generated by BVLC-GoogLeNet and ResNet-152. Metrics are shown below images in form (\textit{PASS}, L$_2$ norm per pixel, L$_\infty$). (a) Fast Gradient Sign (FGS) method with BVLC-GoogLeNet: classified as ``starfish'', failed adversarial generation due to low \textit{PASS} (b) Fast Gradient Value (FGV) approach with BVLC-GoogLeNet: classified as ``loggerhead turtle'', failed adversarial generation due to low \textit{PASS} (c) Hot/Cold approach with the most similar class (HC-1) on BVLC-GoogLeNet: adversarial classified as ``mud turtle'' (d) Hot/Cold approach with the 13th most similar class (HC-13) on BVLC-GoogLeNet: adversarial classified as ``box turtle'' (e) Fast Gradient Sign (FGS) method with ResNet-152: adversarial classified as ``cock'' (f) Fast Gradient Value (FGV) approach with ResNet-152: adversarial classified as ``blowfish'' (g) Hot/Cold approach with the most similar class (HC-1) on ResNet-152: adversarial classified as ``bee'' (h) Hot/Cold approach with the 13th most similar class (HC-13) on ResNet-152: adversarial classified as ``cock''. For better visualization, perturbations are scaled by a factor of 10 with gray=no-change.}
\end{center}
\vspace*{-0.15in}
\end{figure*}

As shown in Fig.~\subref{sfig:a} and Fig.~\subref{face_gen:a}, the adversarial perturbations produced by the FGS method cover almost the entire image, including the ``background''. By contrast, using raw gradient of loss, shown in Fig.~\subref{sfig:b}  and Fig.~\subref{face_gen:b}, generates more focused perturbations that cause less structural damage resulting in higher PASS as we can see in Fig.~\ref{fig:metrics}. Still, FGS and FGV approaches combined produce only two adversarial examples per input image, and it is natural to ask if we can generate more. 

\subsection{Hot/Cold Approach}

\label{sec:hot}
The intuition behind FGS and FGV is to decrease the response to the class of interest by following the direction of the gradient of loss.
To generate hard positives which \textit{augment} overfit decision boundaries between classes, however, it seems natural to consider derivatives with respect to other layers, and generate adversarial examples which change classifications \textit{while moving toward a specific targeted class}.
%To generate even hard positive,, we consider derivatives with respect to other layers and generate adversarial images by both moving away from the current class and while moving in the directions of those other derivatives.  
Our idea is related to \cite{mahendran2014understanding}, which introduced a method called \textit{image inverting}, designed to reconstruct an image which minimizes the loss for given class represented by a \textit{one-hot} feature vector in the penultimate layer. 
We postulated that inverting a one-hot vector -- a vector with one non-zero (hot) element and the remaining elements zeros --  at the penultimate layer would eventually create features in lower layers that are exclusive to the selected \textit{hot class}, and other features responsible for neutralizing other classes represented by zeros in the penultimate layer (\textit{non-hot classes}). We extend that concept by adding the ``cold'' class, to further decrease the role of the current class. Specifically, we craft features for the penultimate layer, the last layer before softmax. 
At that layer each value is still associated with a particular output class, so we can define directions in the input image space that help move toward a target (hot) class while moving away from the original (cold) class. 
%, with features corresponding to other classes effectively zeroed out.
% other features neutralized other features responsible for neutralizing other classes having zeros. We show that those exclusive features of classes can be effectively applied to construct adversarial perturbations.

To formalize our \textit{hot/cold} model, let $h(x) \in {\bf R}^n$ be the features of the neural network's penultimate layer for input image $x$, and let $y$ be the label for $x$. We construct a hot/cold feature vector $\omega_{hc}$ based upon $h(x)$ as follows: First, we define a target class $\tilde{y} \neq y$ as hot, i.e., we add features exclusive to this class $\tilde{y}$ to input image by defining its corresponding value as $|h_{\tilde{y}}(x)|$. Second, we identify class $y$ as cold with a value of $-h_y(x)$ as we intend to step away from it by removing its exclusive features from the input image. Thus the constructed penultimate layer feature vector is
 \vspace*{-.5ex}
\begin{equation}\label{eq:hot_cold}
  \omega_{hc}(x)=\left\{\def\arraystretch{1.2}%
  \begin{array}{@{}c@{\quad}l@{}}
    {|h_j(x)|} & {\mathit{if} \: j=\tilde{y}}\\
    {-h_j(x)} & {\mathit{if} \: j=y}\\
    {0} & {otherwise.}\\
  \end{array}\right.
\end{equation}
To be able to target both similar and dissimilar classes of input image $x$, we use scalar value $|h_j(x)|$ for the hot class. Finally, as described below, we use backpropagation to estimate the image changes needed to move according to our constructed feature vector $\omega_{hc}$ by computing the image $\eta_{hc} = B_l(\omega_{hc})$.
The operator $B_l(\cdot)$ is an approximation to the derivative backpropagated down to the image level. Sec~\ref{sec:hack} explains how this is conducted at intermediate layers. While any positive value for the hot class and any negative value for the cold class have beneficial effects in terms of providing adversarial directions, we find that, in general, values derived from the extracted feature vector of the original image perform very well as they naturally capture the relative scale of the respective class's features.

As we can see in Fig.~\ref{fig:first_gen} for MNIST and in Fig.~\ref{fig:face_gen} for ImageNet, our hot/cold approach provides comparable results (PASS) to the previously described FGV approach in Sec.~\ref{sec:fgv}. However, with this new approach we can explicitly move in the direction of targeted classes obtaining several different adversarial directions for each input image. This greatly increases the adversarial diversity available for training.

\begin{figure}[!tb]
\begin{center}
\centerline{\includegraphics[width=.95\columnwidth]{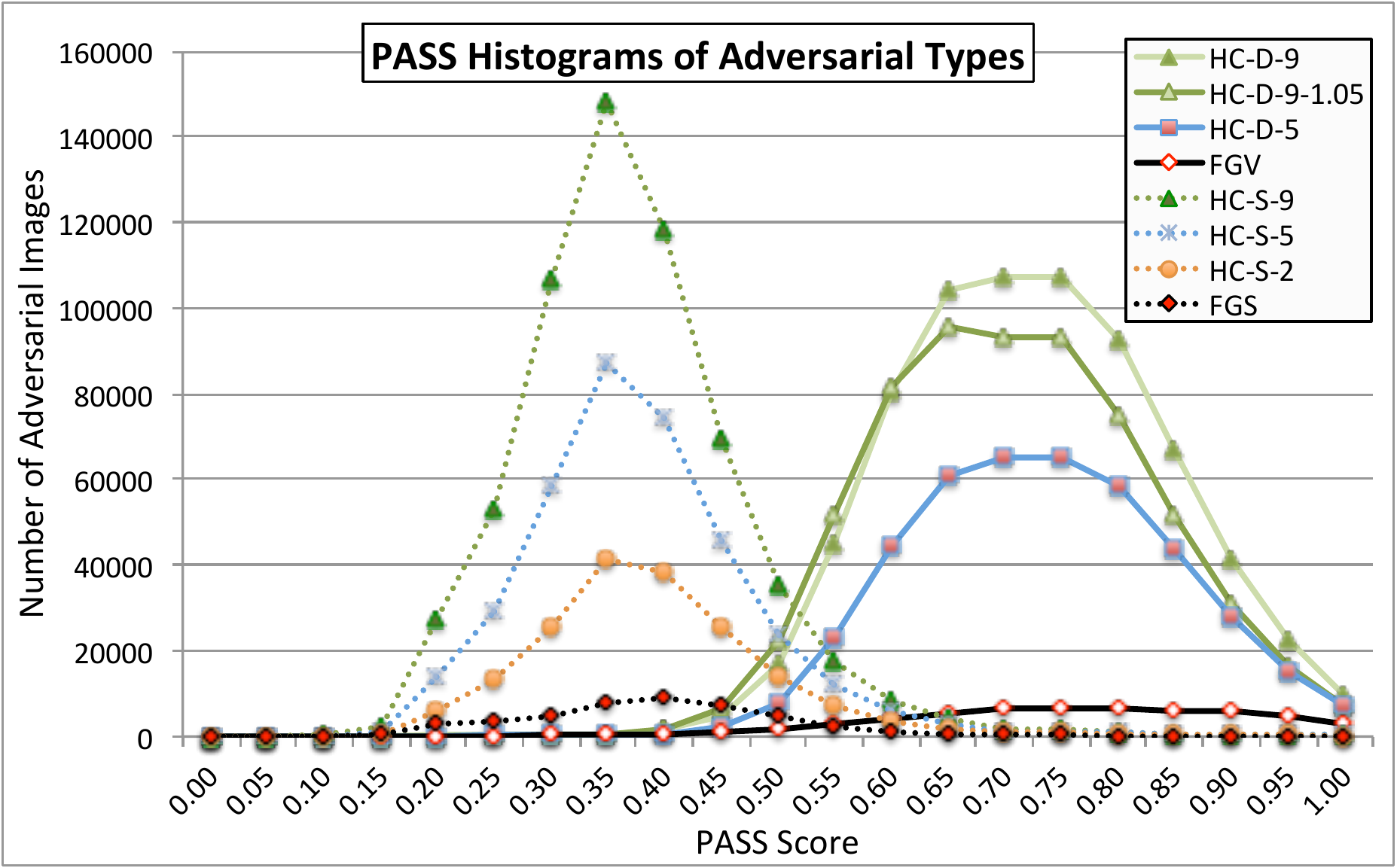}}
\cap{fig:PassPlot}{PASS Scores}{Example showing distributions of PASS scores for different types of adversarial and hard positive LeNet/MNIST images. The HC-D-X distributions use the hot/cold approach (Sec.\ref{sec:hot}) with X closest scoring classes, while HC-S-X use the sign of the derivatives. HC-D-9-Y uses the derivatives scaled by Y beyond the point of minimal adversarial PASS in the direction of the derivatives. FGS is the fast gradient sign method of \protect\cite{goodfellow2014explaining}. These plots demonstrate that using sign instead of raw directions defined by derivatives significantly reduces PASS scores, e.g., compare HC-S-9 and HC-D-9. They also illustrate that even small extensions beyond the minimal adversarial distance reduces PASS as well. Finally, the plots show the sheer increase in number of adversarials over the basic FGS approach.}
\end{center}
\vspace*{-0.25in}
\end{figure}

\subsection{Implementation Details}
\label{sec:hack}

For our adversarial generation implementations and experiments, we use the popular deep learning framework Caffe \cite{jia2014caffe}. Caffe allows us to obtain inner representations of images in neural networks and backpropagate feature representations.
However, Caffe's {\tt backward} method is limited to start with features of the top layer. To perform backpropagation of feature representations at deep(er) layers, there are two options. First, we can create truncated networks, make a targeted layer sit at the top of each, and then use the {\tt backward} method to perform the backpropagations of interest. This approach is cumbersome. The second approach is to modify the {\tt backward} method to directly backpropagate from any specified feature representations. By simply eliminating few lines of code responsible for checking whether the specified starting point is the top layer, we can use a single network for our experiments. This is how we implement the backpropagation operator $B_l(\cdot)$. % at

Our novel approach for obtaining adversarial images calculates feature-derivatives at any layer, and allow us to obtain perturbations which define different directions with respect to a given input image. We can search along those directions for the closest perturbations that cause mislabeling to obtain adversarial images or further extend this search to obtain additional hard positives.
Since we apply a given direction as a perturbation to the input image, by scaling the perturbation by larger and larger values and adding it to the original image, we move the original image farther and farther along that direction. In order to efficiently discover the closest adversarial point in that direction, we apply a line-search technique with increasing step-sizes. To find adversarials, we search for the smallest possible adversarial perturbation in the last section of line-search by applying a binary search. Finally, we would like to emphasize that all images we generate have discrete pixel-values in $[0,255]$.

\begin{figure}[tb]
\begin{center}
\centerline{\includegraphics[width=.95\columnwidth]{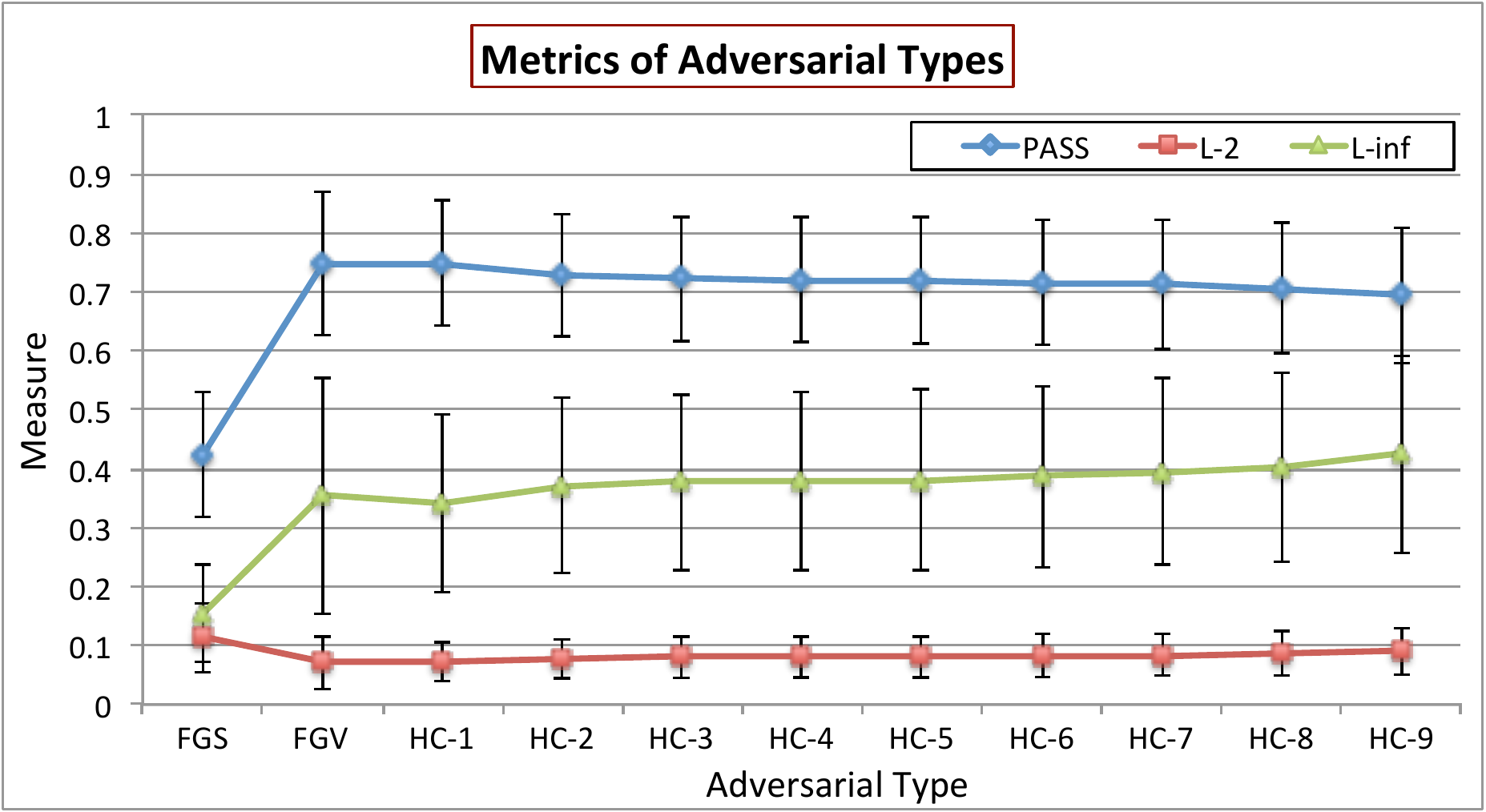}}
%\centerline{\includegraphics[width=.9\columnwidth]{Images/metrics.pdf}}
\cap{fig:metrics}{Metrics}{Example showing various metrics (PASS, $L_2$, $L_\infty$) for different types of MNIST adversarial images generated on three LeNets. FGS, FGV, and HC-X denotes hot/cold approaches targeting the Xth closest scoring classes. While adversarials of FGV and our HC methods have higher $L_\infty$ distances, they also have better PASS scores than FGS. For visualization both $L_2$ and $L_\infty$ distances are scaled so that max distances are 1.}
\vspace*{-0.25in}
\end{center}
\end{figure}

%\vspace{-1em}
\section{Experiments}

The focus of our experiments is threefold. First, we generate adversarial examples on LeNet/MNIST and collect metrics to compare different algorithms in terms of $L_2$ distances, $L_{\infty}$ distances, and PASS. Second, we demonstrate that while PASS can be applied to complex images, ``good'' thresholds for adversarial and hard positive examples vary with the domain. Finally, we demonstrate on LeNet/MNIST that retraining with a diverse set of hard positive images can improve the original network's performance more than just training with adversarials generated via the FGS approach.

\subsection{MNIST - Adversarial Metrics}

%Example showing various metrics (PASS, $L_2$, $L_\infty$) for different types of MNIST adversarial images generated on 3 LeNet networks, described above. HC1 and HC2 denote hot/cold approach to closest and second closest classes, SG-Layer (Goal) stands for source/goal approach. While our new methods have higher $L_\infty$ distances, they also have comparable or better PASS scores than FGS. For better visualization both $L_2$ and $L_\infty$ distances are scaled: possible maximum distances are 1.

Here we display metrics for various types of adversarial examples generated on
LeNets trained with the MNIST dataset. Fig.~\ref{fig:PassPlot} shows the distribution of PASS scores for different types of adversarial images compared with FGS. As we can see in Fig.~\ref{fig:metrics}, $L_2$ and $L_{\infty}$ distances do not map well to perceptual similarity as defined by PASS. For example, while adversarial images generated by the FGS method have noticeably lower $L_{\infty}$ distances than those generated by the FGV method, adversarials of FGV maintain significantly higher PASS. 

\comment{
\subsection{PASS and Face Network Adversarial Images}
While multiple authors have demonstrated adversarial images on ImageNet, we chose to test our model on face images/networks because humans are very sensitive to variations in faces and because fine-grained recognition presents a greater challenge. We show examples generated on a network very similar to those of
\cite{chen2015unconstrained} and \cite{yi2014learning} trained on the CASIA-WebFace dataset \cite{yi2014learning}. Prior work on adversarial images produced
perturbations that appear almost as random noise. As we can see in
Fig.~\ref{fig:face_gen}, the structure of the noise for many of our adversarial examples is concentrated on important facial features. Many of the new techniques, especially using the raw feature-derivatives, produce adversarial images in which the perturbations are visibly structured -- not just mostly on or around the object but also focused on and around the salient features. For example, we can clearly see artifacts of eyes, eyebrows, jawline, and other salient features of the face within our perturbations. 
The lowest level features, e.g., Conv1, are focused on salient edges, either sharpening them by adding the feature-derivative or blurring them by subtracting the feature-derivative.

Despite their complex nature, our perturbations can have very high $L_2$ and $L_\infty$ norms, yet result in images which are still very face like. The similarity scores (PASS) of adversarial CASIA-WebFace images remain high even after applying significant perturbations which we can clearly detect. For example, in Fig.~\subref{face_gen:l} we use the opposite direction of input image's backpropagated Conv1 level feature, which produces a ``blurred'' adversarial image with a relatively high PASS of 0.8928. We also note that perturbations, such as edge-sharpening and blurring, can occur quite naturally, e.g., from image compression or optical distortions. 
We conclude that the adversarial threshold in Eq.~\ref{eq:adv} needs to be defined with respect to the dataset to which it is applied. 
}

\subsection{MNIST - Training with Hard Positives}

We show that fine-tuning with a diverse set of hard positive images
enhances LeNet/MNIST's robustness to such examples and also improves its
accuracy. We trained 3 different LeNet/MNIST networks with the standard 60K
training samples and different random initializations/orderings using the
training parameters distributed with Caffe. While many techniques are known to
improve performance, e.g. batch normalization and different objective functions,
we do not employ those approaches here as the goal is to evaluate the
performance gains on a very standard network and make it easy for others to
replicate/compare.

For each network, we generated adversarial and hard positive images of different types. We then fine-tuned each network three times using different random orderings of the
adversarial training data, giving 9 different networks for testing. Fine-tuning
was for 20K iterations; the base networks were initially trained with 10K
iterations.

To test, we used both the standard MNIST test data of 10K images and a set of 70K held-out adversarial and hard positive images that were not used for training. The 70K held-out set contained a broad mix of adversarial types, 5K each for 14 types: FGS \cite{goodfellow2014explaining}, FGV (cf. Sec.~\ref{sec:fgv}), 9 types of hot/cold approaches (cf. Sec.~\ref{sec:hot}) from HC-1 to HC-9, HC-1 scaled by 1.05, HC-1 scaled by 1.10, and adversarials of the hot/cold approach using signs of derivatives for the closest scoring class. Training used adversarial images from only the associated network on a subset of types. To better test generalization, adversarial and hard positive testing uses samples from all networks, thus the basic LeNets do not show 100\% error.

\begin{figure}[t]
\begin{center}
  \centerline{\includegraphics[width=.95\columnwidth]{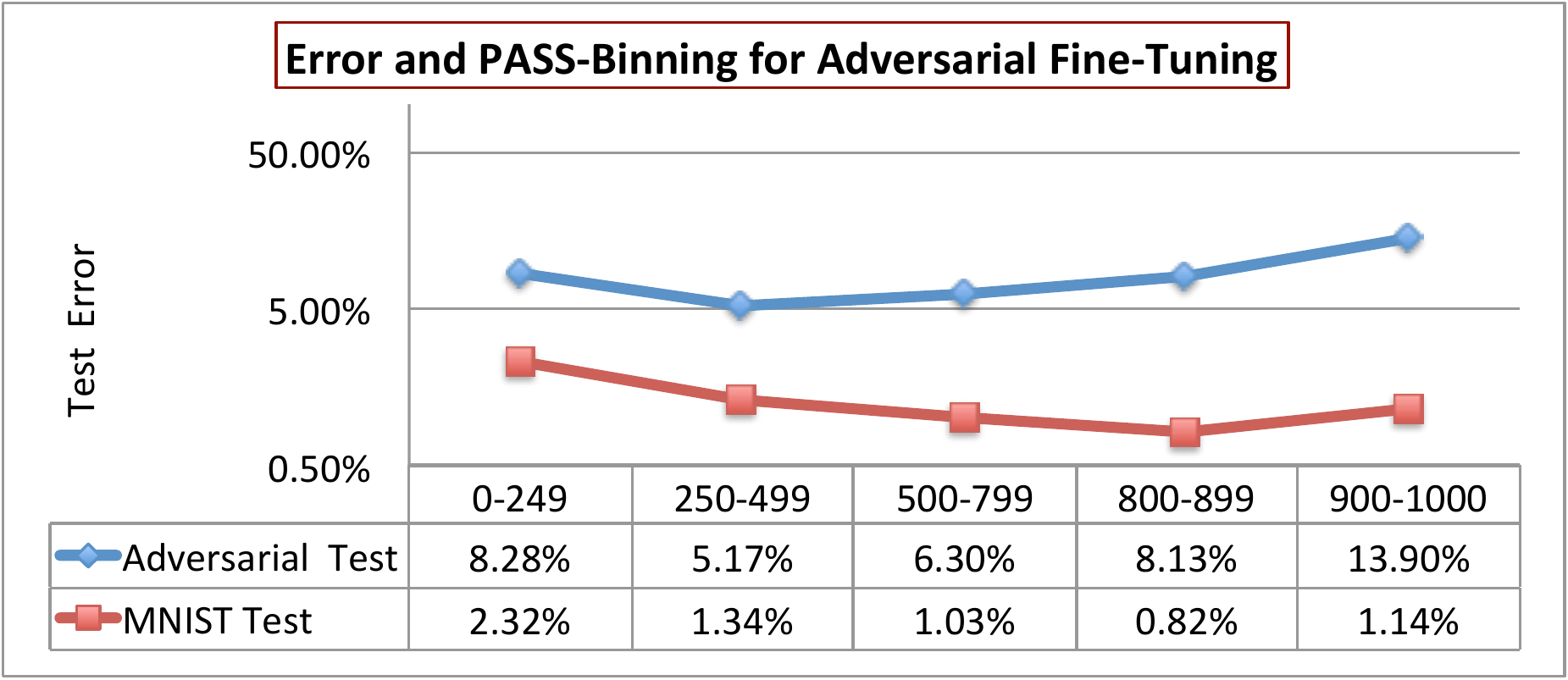}}
  \cap{fig:ErrorPassPlot}{PASS Scores}{Example showing how error on both original and all adversarial MNIST test data varies for networks trained using generated hard positives with different ranges of PASS scores. Note that networks trained on data with the highest PASS scores have higher MNIST test error and much higher adversarial test error. Even though adversarial testing includes data from all ranges, training using data only with PASS scores in the 800-900 range still does well in adversarial testing.}
\end{center}
\vspace*{-0.3in}
\end{figure}

Table~\ref{t:tab1} shows the average accuracies for different test models. These results are consistent both with the hypothesis that increasing diversity improves training and the hypothesis that PASS scores for hard positives should be non-minimal, but not too large. By pushing the boundaries past the minimal adversarial images and closer to the goal class, as graphically depicted as stars in Fig.~\ref{fig:general}, HC-D-9-1.05, hard positives obtianed by scaling perturbations of HC-1 to HC-9 adversarials (HC-D-9), generalizes the boundaries more and thus increases accuracy over HC-D-9 while significantly improving robustness to adversarial images. The results in Fig.~\ref{fig:ErrorPassPlot} demonstrate that training on images of intermediate PASS values produces better results than training using only images with either the highest or lowest PASS scores. To compare with state-of-the-art adversarial training, we note that HC-D-9, HC-D-9-1.05 and even FGV are significantly better than training with FGS images using the approach of \cite{goodfellow2014explaining}. The line with * contains the results reported in \cite{goodfellow2014explaining} using a modified adversarial-enhanced training objective, which is slightly better than training with a finite set of FGS images. This is likely because the objective function can take advantage of a changing adversarial objective during training. 

We also performed comparisons with approaches that use other types of augmented data. We generated 1M images with InfiMNIST \cite{loosli2007training} and trained three networks for 30K iterations. We see that HC-D-9 and HC-D-9-1.05, each of which are trained with approximately 340K adversarial images, are slightly (but not statistically) better than training on 1M images from InfiMNIST. FGV, trained with 37663 images, which is only 3.7\% of the InfiMNIST data, does equally as well. Furthermore, the InfiMNIST trained networks also perform poorly on adversarial examples.

\begin{table}
  \small
  \centering
%  \begin{tabular}{m{2cm}|m{1.5cm}|m{2.5cm}}  \small
      \begin{tabular}{r|c|l}  \small
MNIST  & Adversarial & Method \\ \hline \hline
$.673\%\pm  .035$ &	 $14.85\%$ &        HC-D-9-1.05  \vphantom{\large$|^x)$}    \\ \hline
$.683\%\pm  .057$ &	 $21.10\%$ &        HC-D-9      \vphantom{\large$|^x)$}   \\ \hline
$.697\%\pm  .021$ &	 $48.35\%$ &        InfiMNIST (1M) \vphantom{\large$|^x)$}   \\ \hline
$.697\%\pm  .031$ &	 $29.61\%$ &        FGV    \vphantom{\large$|^x)$}   \\ \hline
%$.713\%\pm  .006$ &	 $15.31\%$ &        HC-D-5-1.05  \vphantom{\large$|^x)$}   \\ \hline
%$.723\%\pm  .01$ &	 $29.47\%$ &        HC-S-2       \vphantom{\large$|^x)$}   \\ \hline
$.723\%\pm  .047$ &	 $21.52\%$ &        HC-D-5      \vphantom{\large$|^x)$}   \\ \hline
$.733\%\pm  .032$ &	 $29.29\%$ &        HC-S-5       \vphantom{\large$|^x)$}   \\ \hline
%$.747\%\pm  .065$ &	 $22.40\%$ &        HC-D-2        \vphantom{\large$|^x)$}   \\ \hline
%$.7533\%\pm  .05$ &	 $8.93\%$ &         HC-D-9-1.1     \\ \hline
%$.782\%$~~~~~~~~~~~~ &          \multicolumn{2}{c}{ \hbox{\cite{goodfellow2014explaining}}}   \vphantom{\large$|^x)$} \\ \hline

$.782\%$ *~~~~~~~~~~      &	 $17.9\%*$ &        Goodfellow et al. \cite{goodfellow2014explaining}\vphantom{\large$|^x)$}   \\ \hline

%$.782\%$~~~~~~~~~~~~ &          \multicolumn{2}{c}{ \hbox{\cite{goodfellow2014explaining}}}   \vphantom{\large$|^x)$} \\ \hline

%$0.783\%\pm  0.13$ &	 $15.96\%$ &        HC-D-2-1.05  \vphantom{\large$|^x)$}   \\ \hline
%$0.783\%\pm  0.13$ &	 $15.96\%$ &        HC-D-2-1.1    \vphantom{\large$|^x)$}   \\ \hline
%$0.790\%\pm  0.01$ &	 $9.69\\%$ &         HC-D-5-1.1    \vphantom{\large$|^x)$}   \\ \hline
$.790\%\pm  .122$ &	 $31.06\%$ &        FGS \vphantom{\large$|^x)$}   \\ \hline
$.937\%\pm.110$& 64.21\% & Basic LeNet \vphantom{\large$|^x)$}   \\ \hline
  \end{tabular}
      \cap{t:tab1}{Error Rates}{Error rates of various adversarial and hard positive trained networks on both MNIST and adversarial test sets. Increasing adversarial diversity clearly improves results. The best performing model, HC-D-9-1.05, using the hot/cold approach with derivatives from all 9 classes to generate hard positives, reduces the MNIST test error 28.18\% over the basic LeNet and 14.81\% over training with fast gradient sign (FGS)~\cite{goodfellow2014explaining}. Note that this model also has the lowest susceptibility to adversarial images with an error rate of 14.85\%.}
\end{table}

\subsection{ImageNet - Adversarial Training}

We conducted preliminary experiments using our diverse adversarial approach to enhance the performance of BVLC-GoogLeNet on the ImageNet dataset. Instead of training a new network from scratch, we show that we can improve the existing BVLC-GoogLeNet model by fine-tuning it with diverse adversarial images. Because our focus is on using hard positives to enhance learning, we take only the center crop (which is used by the pre-trained BVLC-GoogLeNet). We started by taking 15 correctly classified images per class, and generated 20 different HC-types -- from the most similar (HC-1) to twentieth most similar (HC-20) class as hot, yielding 300K images in total. We filtered out radical perturbations, keeping only the adversarial images with \textit{PASS} scores higher than 0.99. This yields approximately 250K hard positives for training. We then mixed the 250K adversarial images with the original approximately 1.28M center crops of the ImageNet training set. Our fine-tuning procedure relies on original hyperparameters distributed with the BVLC-GoogLeNet model in the Caffe Model Zoo. We fine-tuned the original BVLC-GoogLeNet model for 500K iterations with batch size of 80, i.e. about 25 epochs. We tested on center crops of the ImageNet validation set and report top-1 and top-5 error rates. The top-1 error is 30.552\%, and top-5 error rate is 10.604\%. While this performance is not state-of-the-art for ImageNet, our improvement in top-1 error rate is 2.07e-4\% per added image, and our top-5 gain is 1.01e-4\% per added image. As computed from the data in \cite{szegedy2015going}, using 10 crops decreased their top-5 error rate with 7.67e-6\% per added image, while the use of 144 crops reduced top-5 error by 1.26e-06\% per added image. Therefore, using our hard positives in training yields greater improvement for each additional image.
%Future work will explore similar fine-tuning approach on state-of-the-art ResidualNets. 

\section{Conclusion}

In this paper, we have introduced a new perceptual measure for \textit{adversarial images}, which allows us to explicitly quantify the constraint that perturbations must be imperceptible in order to form adversarial images. We have also introduced novel ways of generating diverse adversarial images, and have shown that amplifying adversarial images to create additional \textit{hard positives} can be used to augment training datasets and further enhance accuracies over training with adversarial images alone. Instead of looking like random noise, many of our diverse adversarial perturbations exhibit strong structural artifacts, which are far more likely to occur naturally than perturbations of former adversarial generation methods that generally look like random noise.

Although we could not achieve state-of-the-art performance on MNIST or ImageNet, our results on MNIST are substantially better than those of any prior work leveraging adversarial training or input perturbations on LeNet/MNIST. Our approach also uses far fewer training images and improves robustness to adversarial examples. On ImageNet, we show compelling evidence to suggest that using hard positive images in the training set offers substantially greater performance improvement per image than using translated crops.
%Future work will explore combining adversarials with crops, and will be applied to fine-tune the new state-of-the-art ResidualNet \cite{he2015deep}.
We have shown adversarial images generated on the state-of-the-art ResNet-152, which suggests that our approach can likely be applied to improve the state of the art.

{\small
\bibliographystyle{ieee}
\bibliography{paper}
}

\end{document}